\documentclass[10pt,twocolumn,letterpaper]{article}

\usepackage{ijcb}
\usepackage{times}
\usepackage{subcaption}
\usepackage{url}
\usepackage{epsfig}
\usepackage{graphicx}
\usepackage{amsmath}
\usepackage{amssymb}
\usepackage[flushleft]{threeparttable}
\usepackage{collcell}
\makeatletter
\newcolumntype{n}{>{\collectcell\@gobble}c<{\endcollectcell\hskip-2\tabcolsep}}
\makeatother
\usepackage{enumitem}
\usepackage{array, makecell} %
\usepackage{tabularx,booktabs}

\ijcbfinalcopy 

\begin{document}

\title{OCFR 2022: Competition on Occluded Face Recognition From Synthetically Generated Structure-Aware Occlusions}

{
\author{ 
Pedro C. Neto$^{1,2,+}$,
Fadi Boutros$^{3,4,+}$,
Joao Ribeiro Pinto$^{2,+}$, 
Naser Damer$^{3,4,+}$,\\
Ana F. Sequeira$^{1,+}$, 
Jaime S. Cardoso$^{2,1,+}$, \vspace{2mm} \\ 
Messaoud Bengherabi$^{5,*}$,
Abderaouf Bousnat$^{6,*}$,
Sana Boucheta$^{6,*}$,
Nesrine Hebbadj$^{6,*}$,\\
Mustafa Ekrem Erak{\i}n$^{7,*}$,
U\u{g}ur Demir$^{7,*}$, 
Haz{\i}m Kemal Ekenel$^{7,*}$,\\
Pedro Beber de Queiroz Vidal$^{8,*}$, 
David Menotti$^{8,*}$,
\\
$^{1}$INESC TEC, Portugal -
$^{2}$University of Porto, Portugal\\
$^{3}$Fraunhofer Institute for Computer Graphics Research IGD, Germany -
$^{4}$TU Darmstadt, Germany\\
$^{5}$CDTA Centre de Développement des Technologies Avancées, Algeria\\
$^{6}$ESI Ecole Nationale Supérieure d'Informatique, Algeria\\
$^{7}$SiMiT Lab, Istanbul Technical University, Turkey \\
$^{8}$Federal University of Parana, Brazil \\
$^{+}$Competition organizer. $^{*}$Competition participant.\\
Email: {pedro.d.carneiro@inesctec.pt}
\vspace{-4mm}
}
}
%

\maketitle
\thispagestyle{empty}

\begin{abstract}

{This work summarizes the IJCB Occluded Face Recognition Competition 2022 (IJCB-OCFR-2022) embraced by the 2022 International Joint Conference on Biometrics (IJCB 2022). OCFR-2022 attracted a total of 3 participating teams, from academia. Eventually, six valid submissions were submitted and then evaluated by the organizers. The competition was held to address the challenge of face recognition in the presence of severe face occlusions. The participants were free to use any training data and the testing data was built by the organisers by synthetically occluding parts of the face images using a well-known dataset. The submitted solutions presented innovations and performed very competitively with the considered baseline. A major output of this competition is a challenging, realistic, and diverse, and publicly available occluded face recognition benchmark with well defined evaluation protocols. }
\end{abstract}

\vspace{-4mm}
\section{Introduction} 

Over the past two years, the COVID-19 pandemic has intensified the occurrence of a particular type of face occlusion, the facial mask. Due to sanitary concerns, the majority of the countries introduced an obligation to wear masks in closed or crowded environments. As expected, it led to a significant performance drop in face recognition (FR) scenarios~\cite{DBLP:conf/biosig/DamerGCBKK20,gomez2021biometrics,DBLP:journals/iet-bmt/DamerBSKK21}, as shown on the latest FR vendor test conducted by NIST~\cite{frvt6a}. Similarly, Damer~\textit{et al.}~\cite{damer2022masked} have shown that the performance of humans {in recognising faces} is also negatively affected by the presence of a face mask. These findings led to a new direction in the FR {research}. Hence, an additional effort was made to solve the masked face recognition {(MFR)} problem, which benefit from contributions in the form of case-study papers~\cite{fu2021effect,DBLP:journals/pr/FangDKK22,wang2021survey,seneviratne2022does}, methods~\cite{neto2021focusface,huber2021mask,neto2021my,DBLP:journals/pr/BoutrosDKK22} and competitions~\cite{boutros2021mfr,deng2021masked}. Moreover, {the effect of masks} were not exclusively linked to recognition {and studies can be found} on their effect on face presentation attack detection methods~\cite{DBLP:conf/fgr/FangBKD21,DBLP:journals/pr/FangDKK22}. The volume of new research led to a second part of the NIST FR vendor test, which evaluated the performance of FR algorithms designed post-covid~\cite{frvt6b}. 

While MFR is a special case of a broader problem known as occluded face recognition (OFR), the research output on the latter was not significantly affected by the pandemic. Surprisingly, the vast majority of the solutions proposed to solve MFR were not inspired by previous OFR research~\cite{Neto2022beyondMasks}. Thus, there is an opportunity to work on the broader problem with solutions inspired by the more specific one. There are a few works on OFR post-pandemic. Erakiotan~\textit{et al.}~\cite{erakiotan2021recognizing} proposed a new dataset composed of real faces occluded by sunglasses or face masks. Qiu~\textit{et al.}~\cite{qiu2021end2end} proposed an improved framework over the previous state-of-the-art method by Song~\textit{et al.}~\cite{song2019occlusion}. Their method, FROM, decreases the training and testing complexity by removing the need for two different inference stages in both training and testing. Both Huang~\textit{et al.}~\cite{Huang2022jointsegmentation} and Zeng~\textit{et al.}~\cite{zeng2021occlusion} propose architectures that, due to a segmentation module, are occlusion-aware. He~\textit{et al.}~\cite{he2021locality} studied a variation of dropout to maximise the performance under occlusions. Finally, Zeng~\textit{et al.}~\cite{zeng2021survey} conducted an extensive research on the techniques used for OFR. 

The majority of the works previously mentioned are evaluated on small scale datasets composed of real occlusions or on larger datasets with synthetic occlusions arbitrarily added to face images. These two different approaches for model evaluation produce different results. This finding is stated by Jiang~\textit{et al.}~\cite{jiang2022new} after performing the necessary comparative studies. One key highlight from their work is that current models are being tested on strange and abnormal synthetic occlusions, such as blocks. The insertion of a structure to the occlusions added on a synthetic dataset has already been intitially approached by Huang~\textit{et al.}~\cite{huang2021face}. However, their approach lacks variety and does not attempt to combine multiple occlusions on different structural areas. Thus, there is a need for more realistic face occlusions data that allow the development of methods that are more robust in real world applications.

One difference between OFR and MFR research is that the latter usually aims to be an extension of FR, in the sense, that several of the proposed model introduce little to no overhead or complexity. Driven by an increased deployment on consumer and embedded devices, there is an interest in lightweight FR approaches, as shown in 2019 ICCV competition~\cite{deng2019lightweight}. Some models from that competition and on follow-up research have shown an interesting trade-off between complexity and performance~\cite{chen2018mobilefacenets,boutros2021mixfacenets,DBLP:journals/access/BoutrosSKDKK22}. As stated, research on MFR runs in many cases on top of existing models, making it model-agnostic, such as in \cite{DBLP:journals/pr/BoutrosDKK22}. Hence, it can also benefit from these lightweight models, whereas OFR is mostly application dependent.

Inspired by: (a) the effective research that has been conducted over the past two years on MFR; 
(b) the larger information availability variation in OFR in comparison to MFR; 
(c) the need to revitalise the field with novel research in the direction of improving OFR performance; and (d) the need for a challenging benchmark truly inspired by real occlusions and a notion of structure on where synthetic occlusions are added, we conducted the IJCB Occluded Face Recognition Competition 2022 (IJCB-OCFR-2022). The final participation toll comprises 3 teams, which have submitted two solutions each, totalling 6 valid solutions. The solutions were evaluated on our novel benchmark including seven databases comprising occlusions placed over four different structural areas, which can also be combined into stronger and broader occlusions. This paper summarises this competition, the submitted solutions, the novel benchmark and the achieved results in terms of occluded vs occluded face verification accuracy, and occluded vs non-occluded face verification accuracy, as well as the compactness of the recognition models. 
In the following sections, we introduce the new benchmark used for the competition and the associated evaluation criteria, in Section~\ref{sec:data_evaluation}. In Section~\ref{sec:submitted}, we describe the six valid solutions submitted by three teams. The results achieved are discussed in Section~\ref{sec:results}. Finally, a brief conclusion is provided in Section~\ref{sec:conclusion}. 
  
\section{Database and evaluation criteria}
\label{sec:data_evaluation}

\subsection{OCFR-2022 evaluation benchmark}

The evaluation dataset is built from Labeled Faces in the Wild (LFW) dataset~\cite{LFWTech} by synthetically occluding parts of the face images. This dataset is referred to as occluded face recognition competition data (OCFR-2022) and it is used to evaluate the verification performance of the submitted solutions.
The OCFR-2022 evaluation benchmark will be made publicly available~\footnote{\url{https://github.com/NetoPedro/OCFR-2022}} as a batch implementation that would take the LFW as an input and produce the OCFR-2022 evaluation benchmark.

The synthetic occlusions are added in a realistic manner based on three assumptions: 1) there is an inherent structure to the occlusions that are seen in real datasets; 2) some occlusions are only fit to certain spatial areas in the face; and 3) it is possible to have a combination of occlusions.

\begin{table}[!h]
 \caption{Occlusions}
\label{table_occlusions_types}
\centering
\begin{tabular}{cl}
\hline
Type     &  Occlusions \\ 
\hline
 U & Occludes from the nose to the forehead  \\
 L & Occludes from the chin to the nose  \\ 
 T & Occludes part of the forehead \\ 
 E & Occludes the eyes  \\ 
\bottomrule
\end{tabular}
\end{table}

\begin{figure}[h!]
    \centering
    \begin{subfigure}[b]{0.155\linewidth}
       \includegraphics[width = \linewidth]{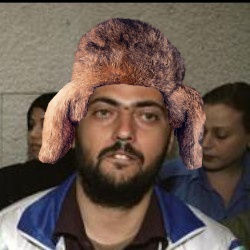}
       \label{unmaskedgood}
  \end{subfigure}
  \begin{subfigure}[b]{0.155\linewidth}
       \includegraphics[width = \linewidth]{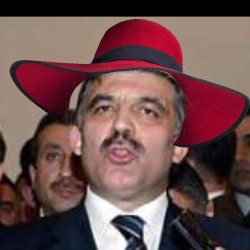}
       \label{maskedgood}
  \end{subfigure}
  \begin{subfigure}[b]{0.155\linewidth}
       \includegraphics[width = \linewidth]{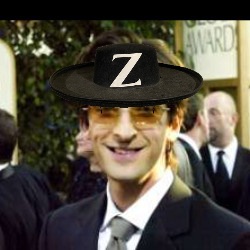}
       \label{unmaskedbad}
  \end{subfigure}
  \begin{subfigure}[b]{0.155\linewidth}
       \includegraphics[width = \linewidth]{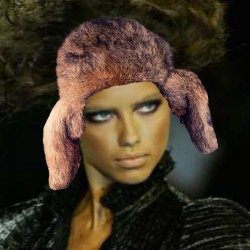}
       \label{maskedbad}
  \end{subfigure}
  \begin{subfigure}[b]{0.155\linewidth}
       \includegraphics[width = \linewidth]{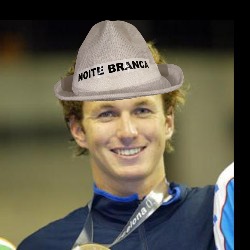}
       \label{maskedbad}
  \end{subfigure}
  \begin{subfigure}[b]{0.155\linewidth}
       \includegraphics[width = \linewidth]{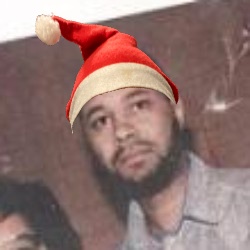}
       \label{maskedbad}
  \end{subfigure}
    
    \begin{subfigure}[b]{0.155\linewidth}
       \includegraphics[width = \linewidth]{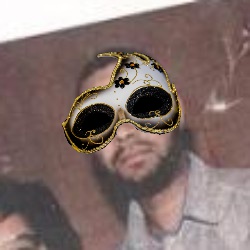}
       \label{unmaskedgood}
  \end{subfigure}
  \begin{subfigure}[b]{0.155\linewidth}
       \includegraphics[width = \linewidth]{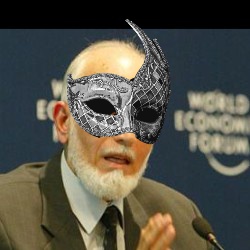}
       \label{maskedgood}
  \end{subfigure}
  \begin{subfigure}[b]{0.155\linewidth}
       \includegraphics[width = \linewidth]{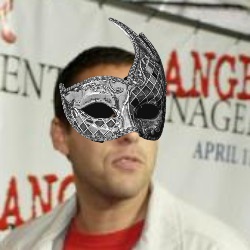}
       \label{unmaskedbad}
  \end{subfigure}
  \begin{subfigure}[b]{0.155\linewidth}
       \includegraphics[width = \linewidth]{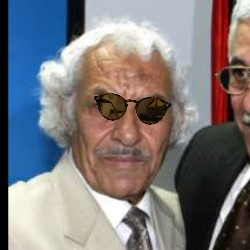}
       \label{maskedbad}
  \end{subfigure}
  \begin{subfigure}[b]{0.155\linewidth}
       \includegraphics[width = \linewidth]{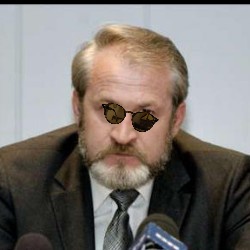}
       \label{maskedbad}
  \end{subfigure}
  \begin{subfigure}[b]{0.155\linewidth}
       \includegraphics[width = \linewidth]{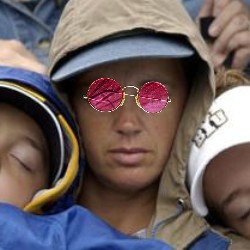}
       \label{maskedbad}
  \end{subfigure}
    
  \begin{subfigure}[b]{0.155\linewidth}
       \includegraphics[width = \linewidth]{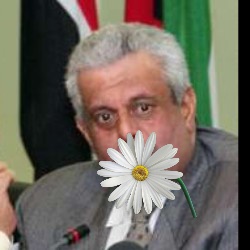}
       \label{unmaskedgood}
  \end{subfigure}
  \begin{subfigure}[b]{0.155\linewidth}
       \includegraphics[width = \linewidth]{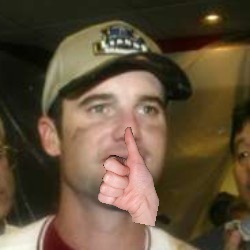}
       \label{maskedgood}
  \end{subfigure}
  \begin{subfigure}[b]{0.155\linewidth}
       \includegraphics[width = \linewidth]{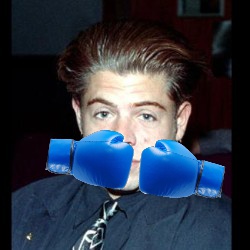}
       \label{unmaskedbad}
  \end{subfigure}
  \begin{subfigure}[b]{0.155\linewidth}
       \includegraphics[width = \linewidth]{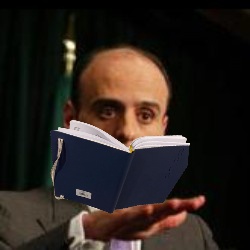}
       \label{maskedbad}
  \end{subfigure}
  \begin{subfigure}[b]{0.155\linewidth}
       \includegraphics[width = \linewidth]{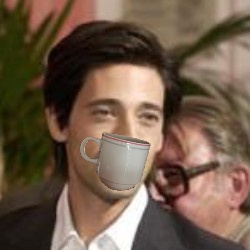}
       \label{maskedbad}
  \end{subfigure}
  \begin{subfigure}[b]{0.155\linewidth}
       \includegraphics[width = \linewidth]{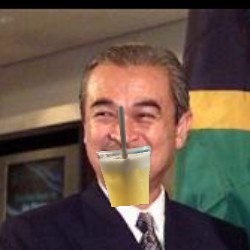}
       \label{maskedbad}
  \end{subfigure}
  \caption{Examples of occluded face images of the OCFR-2022 benchmark comprising top of head (e.g. hats), upper (e.g. carnival masks), eye-based (e.g. sunglasses), and lower (e.g. mouth occlusions such as books, masks) face occlusions. Original non-occluded images are from the Labeled Faces in the Wild (LFW) dataset~\cite{LFWTech}. }
  \label{fig:ex_dataset} 
\end{figure}

For the first assumption, we considered the following structure for introducing the occlusions: top of the head (T), eye-based (E), upper (U) and lower (L) face, as seen in Table~\ref{table_occlusions_types}. These types of occlusions cover different parts of the face. Thus, as stated in the second assumption, each has specific occluders, for instance, sunglasses are not expected to be placed on top of the mouth nor hats are expected to be on the lower part of the face. Some of these occlusions can be seen in Figure~\ref{fig:ex_dataset}. To comply with this assumption, it was necessary to individually and manually annotate all the occluders with the expected position of at least two of the five face landmarks. In other words, for sunglasses, it is necessary to annotate the position of the left eye in the left lens and of the right eye in the right lens. It allows the alignment with the detected landmarks in the original image. We further add some noise drawn from a uniform distribution to the landmarks for variability purposes.

\begin{figure}[h!]
    \centering
    \begin{subfigure}[b]{0.32\linewidth}
       \includegraphics[width = \linewidth]{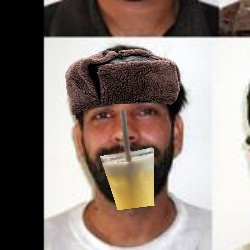}
       \label{unmaskedgood}
  \end{subfigure}
  \begin{subfigure}[b]{0.32\linewidth}
       \includegraphics[width = \linewidth]{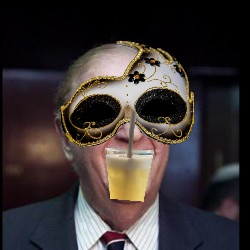}
       \label{maskedgood}
  \end{subfigure}
  \begin{subfigure}[b]{0.32\linewidth}
       \includegraphics[width = \linewidth]{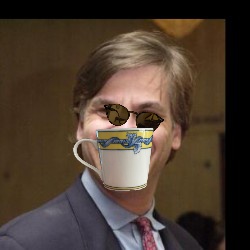}
       \label{unmaskedbad}
  \end{subfigure}

  \caption{Examples of occluded face images comprising, from left to right, a combination of top of head and lower face (T+L); upper and lower face (U+L); and eye-based and lower face occlusions(E+L). Original non-occluded images are from the Labeled Faces in the Wild (LFW) dataset~\cite{LFWTech}. }
  \label{fig:ex_dataset2} 
\end{figure}

Finally, for the third, we propose to combine several occlusions of different types. For instance, an E and an L occlusion can be combined to create face which is strongly occluded, as seen in Figure~\ref{fig:ex_dataset2}.

Despite the liberty to create different configurations through the combination of several occlusions, some combinations are not allowed. For example, adding glasses (occlusion of type E) on top of a carnival mask (occlusion of type U) is not allowed, as they occlude a similar face area. The combinations that were used to create the seven occluded benchmark protocols are in Table~\ref{table_protocols}. Our experiments with the baseline have shown that occlusions of type T were not challenging when used alone. Hence, they were never used independently.

\begin{table}[!h]
 \caption{Evaluation protocols}
\label{table_protocols}
\centering
\begin{tabular}{lc}
\hline
Protocol     &  Occlusions \\ 
\hline
 \#0 & No occlusion\\
 \#1 & U, E or L   \\
 \#2 & L + [UET] or T + [UE]   \\ 
 \#3 & L + [UET], T + [UE] or \#1 \\ 
 \#4 & L + [ET], T + [UE] or \#1 \\
 \#5 & L + T + [UE]  \\ 
 \#6 & L + T + [UE] or \#2 \\ 
 \#7 & L + T + [UE] or \#3 \\ 
  

\bottomrule
\end{tabular}
\end{table}

The LFW dataset~\cite{LFWTech}, used to create the evaluation benchmark, was released in 2007, and since, has become a popular benchmark to compare FR models. 
It is composed of 13,233 face images of 5,749 identities in unconstrained environments. The number of comparison pairs in protocol~\cite{LFWTech} of LFW is 6000 (3000 genuine and 3000 imposter comparisons).
Nonetheless, our approach to generate synthetic occluded images can be applied in any FR dataset without restrictions. 
Based on the occlusion protocols (Table \ref{table_protocols}) and 6000 comparisons pairs provided by the LFW official evaluation protocol~\cite{LFWTech}, we constructed 14 different versions of the dataset. For each occlusion protocols (Table \ref{table_protocols}), we created two versions of occluded LFW. The probes are always occluded. The references are only occluded in the second version.


\begin{table*}[!h]
 \caption{Participants}
\label{table_teams}
\centering
\resizebox{\textwidth}{!}{
\begin{tabular}{cl}
\hline
Team name     &  Affiliation \\ 
\hline
BIOSMC-CDTA & CDTA- Centre de Développement des Technologies Avancées; ESI - Ecole Nationale Supérieure d'Informatique\\
SiMiT Lab & Smart Interaction and Machine Intelligence Lab, Istanbul Technical University, Turkey  \\
UFPR-UNICO & Federal University of Parana, Brazil  \\ 
\bottomrule
\end{tabular}
}
\end{table*}

\subsection{Evaluation criteria}

For the evaluation of the models, we compare their performance on seven different sets of the OCFR-2022 benchmark, which comprise different combinations of occlusions. Besides these datasets, we also include an evaluation of the original LFW dataset~\cite{LFWTech} (without additional occlusions) and a comparison of the compactness of the models, which are not used to rank the solutions for the competition. 
For ranking purposes, the models will be evaluated with non-occluded references and occluded probes. We consider this scenario to be the most likely to occur in the real world, and it will be noted as BLR-OP. We further report the performance of occluded reference and occluded probes on two of the seven datasets. This scenario is noted as OR-OP. 

The verification performance is reported as the false non-match rate (FNMR) at two different operation points, FMR100 and FMR1000, which are the lowest FNMR for a false match rate (FMR) $<$ 1.0\% and $<$0.1\%, respectively. 
For each evaluation setups, the submitted solutions are ranked based FMR100.
Finally, we further report the Equal Error Rate (EER) for each evaluation setups.

The final team's ranking is based on the average of the ranking attained on the seven protocols. Hence, a smaller value indicates a superior solution in terms of ranking order. 

\paragraph{Baseline Methods}
The state-of-the-art FR performance has grown significantly over time. Hence, to better understand and study the performance of the submitted algorithms, it is necessary to compare them with a state-of-the-art FR baseline approach. The selected baseline is the ElasticFace~\cite{boutros2022elasticface}, which achieved state-of-the-art performance on several FR benchmarks. In this work, we use the ElasticFace-Arc+~\cite{boutros2022elasticface} that avoids the assignment of fixed margin value in margin-penalty softmax loss through a flexible penalty margin. The backbone of the ElasticFace is a ResNet-100~\cite{he2016deep} trained on a refined version of the MS-Celeb-1M dataset~\cite{guo2016ms} (MS1MV2). We used a pretrained version of ElasticFace known as ElasticFace-Arc+ provided by the authors~\cite{boutros2022elasticface}\footnote{\url{https://github.com/fdbtrs/ElasticFace}}. We compare the submitted solutions with two versions of the ElasticFace Arc+: \textbf{Baseline UB} and \textbf{Baseline}. The first has access to the landmarks detected from the non-occluded version of the data, which are used to align the dataset. This version is considered an upper bound of the baseline performance, hence, it is known as \textbf{Baseline UB}. The other version only has access to the bounding boxes, which are the only inputs given in the competition. This version, known as \textbf{Baseline}, does not perform alignment nor landmark detection, and it is a less challenging version of the baseline ran on the same conditions of the submitted solutions. One must keep in mind that the baseline was trained to expect an aligned face image. 

\subsection{Submission and evaluation process}

Each submission was requested as a Win32 or Linux console application. However, this was later extended to allow python scripts. A template script and the virtual environment requirements were shared with the participants. The solution should accept the evaluation-list (text file), bounding-boxes (text file) and an output path as parameters. The first input contains the pairs of paths for both the reference and the probe image. The bounding-boxes file contains the bounding boxes detected on the original LFW dataset using the MTCNN algorithm~\cite{zhang2016joint}. We note that, for realistic evaluation purposes, we did not provide the detected landmarks. The output of the solution is a text file, which contains the comparison scores between each pair in the evaluation list. Additionally, participants who do landmark detection and alignment were asked to provide feedback regarding the success of the detection. 


\begin{table*}[!h]
 \caption{Face verification results on three different protocols: protocol 0, protocol 1 and protocol 2 (in the BLR-OP setting). The values of EER, FMR100, FMR1000 are displayed in percentage (\%). The rank orders the solutions from the best to the worst within that protocol. }
\label{results_table_1}
\centering
\resizebox{\textwidth}{!}{
\begin{tabular}{|l|cccc|nnncccc|cccc|}
\hline
  & \multicolumn{4}{|c}{Protocol 0} & \multicolumn{7}{|c}{Protocol 1} &\multicolumn{4}{|c|}{Protocol 2}    \\ 
\cmidrule{1-16} Method &  EER & FMR100 & FMR1000 & Rank  & GMean & IMean & AUC & EER & FMR100 & FMR1000 & Rank& EER & FMR100 & FMR1000 & Rank\\ 
\hline
\hline
 Baseline UB& 0.333 & 0.267 & 0.333 & - & 0.522 & 0.002 & 0.997 & 1.633 & 1.967 & 4.700 & - & 13.000 & 20.567 & 24.100 & -\\
 \hline
 Baseline & 2.500  & 3.433 & 8.633 & - & 0.361 & 0.029 & 0.967 & 8.600  & 20.533 & 37.300 & -& 22.967 & 46.500 & 65.467& -\\
  \hline
  \hline
 SMT-OCFR1 &  0.333 & 0.200 & 0.400 & 1 & 0.394 & 0.757 & 0.956 & 9.667 & 17.100 & 27.033 & 3& 25.633 & 50.033 & 58.900& 4\\
 \hline
 SMT-OCFR2 &  0.333 & 0.300 & 0.667 & 2 & 0.359 & 0.756 & 0.953 & 10.400 & 20.467 & 33.500 & 4& 23.167 & 49.233 & 61.433& 3\\
 \hline
 AFOIRNet\_1 & 15.866 & 50.800 & 67.000 & 5& 0.706 & 0.638 & 0.767 & 31.200 & 68.800 & 85.333 & 5& 35.833 & 84.400 & 93.300& 5\\
 \hline
 AFOIRNet\_2 & 18.466 & 56.167 & 73.400 & 6& 0.673 & 0.610 & 0.738 & 33.400 & 72.400 & 84.733 & 6& 37.467 & 84.500 & 95.033& 6\\
\hline
 AdaFace4M & 1.333 & 1.367 & 2.267 & 4& 0.384 & 0.014 & 0.989 & 4.267 & 8.433 & 19.133 & 2& 15.667 & 29.900 & 40.967 & 2\\
\hline
 AdaFace12M & 1.033 & 1.067 & 1.900 & 3& 0.376 & 0.009 & 0.991 & 3.867 & 6.300 & 15.400 & 1& 15.700 & 27.267 & 38.500& 1\\

\hline
\end{tabular}
}
\end{table*}

\section{Participants and submitted solutions}
\label{sec:submitted}

{The competition aimed to include a diverse set of participants with a significant geographic variation. The call for participation was shared on the International Joint Conference on Biometrics (IJCB 2022) website; on the website dedicated to the competition~\footnote{\url{https://vcmi.inesctec.pt/OCFR2022/}}; on public computer vision mailing lists (e.g. CVML e-mailing list); and on the organisers private e-mailing lists. The call attracted 8 teams, from which, 3 submitted valid solutions. Each team was authorised to submit two solutions. In total there are 6 valid submissions, all from teams with academic affiliations. }

\subsection{BIOSMC-CDTA Team} 

{The methods submitted by the  BIOSM-CDTA  team: \textbf{AdaFace4M} and \textbf{AdaFace12M} are based on the recently released deep models
based on Quality Adaptive Margin loss dubbed AdaFace~\cite{AdaFace}. These methods do not perform face alignment and use the bounding box provided.
\\
Considering that AdaFace improves
largely the FR performance on low
quality academic datasets~\cite{AdaFace}, the aim of the proposed methods is to investigate the
performance of AdaFace on the challenging scenario presenting low quality
occluded faces.
The two proposed solutions (AdaFace4M and AdaFace12M) are trained using million-scale subsets of the cleaned version of the ultra-large-scale face benchmark consisting of 4Midentities/260M faces (WebFace260M)~\cite{WebFace260M}. AdaFace12M is trained on a larger subset of 12M of the curated WebFace42M. The training procedure can be found in~\cite{AdaFace} and the implementation is provided under the MIT license by the authors of~\cite{AdaFace}.
While AdaFace competes with the state of the art angular margin losses on high quality datasets, it outperforms many competing losses on low quality ones, including some quality aware FR systems~\cite{AdaFace}. Another motivation for proposing AdaFace to tackle the occluded FR challenge is the fact that non-severe occlusions can result in hard to recognise low quality face images (still recognisable), which is the one of the advantages of AdaFace. The submitting team states that to the best of their knowledge, AdaFace has not been investigated for the task of occluded FR. The input size of the presented solution is 112×112 and the output feature embedding size is 512-D.}

\subsection{SiMiT Lab Team} 

{Both submissions of the SiMiT Lab Team: \textbf{SMT-OCFR1} and \textbf{SMT-OCFR2} are based on a fine-tuned pre-trained model based on the  Arcface loss~\cite{deng2018arcface}. The pre-trained model used in the competition baseline model (LResNet100E-IR). The LResNet100E-IR model was originally trained on MS-Celeb-1M dataset~\cite{guo2016ms} (MS1MV2), and fine-tuned on real world occlusion datasets. For both upper and lower face occlusions, the authors gathered the Real World Occluded Faces (ROF)~\cite{erakiotan2021recognizing} dataset that consists of sunglasses, masks and neutral face images. As an addition to the lower face occlusion dataset, the ROF dataset was combined with the MFR2~\cite{anwar2020masked} dataset. The proposed methods perform face detection and alignment. MTCNN~\cite{zhang2016joint} model is used to obtain the bounding box and 5 face landmarks. Using those, similarity transform is applied. All the face detection and alignment algorithms are obtained from the original Insightface library\footnote{\url{https://github.com/deepinsight/insightface}}.}

\paragraph{SMT-OCFR1:}The objective is to detect if the occlusion is upper or lower face. Depending on the type of occlusion, is performed FR using models, trained using the Arcface loss~\cite{deng2018arcface}, that are either trained on lower or upper face occlusion datasets. Occlusion detection network is a simple CNN model and trained on ROF sunglasses and ROF masks datasets. Upper face occlusion Arcface model is fine-tuned on ROF sunglasses dataset. Lower face occlusion Arcface model is fine-tuned on ROF masks and MFR2 datasets.

\paragraph{SMT-OCFR2:}{The objective is fine-tuning a pre-trained model, trained using the Arcface~\cite{deng2018arcface} loss, using lower and upper face occlusion datasets together to make a more robust model. Model is fine-tuned on ROF sunglasses, ROF masks, and MFR2 datasets. }

\subsection{UFPR-UNICO Team} 

{The methods submitted by the UFPR-UNICO team: \textbf{AFOIRNet\_1} and \textbf{AFOIRNet\_2} are based on fine-tuning the model proposed by~\cite{huber2021mask}, with non occluded and synthetic occluded images as training data, with the objective of making a model that can discriminate occluded faces. These models do not perform face alignment and use the bounding box provided.}

{To create the two versions of the ``Aspirant facial occlusion invariant recognition network'' (AFOIRNet), the authors opted to generate two occluded versions of the CASIA-WebFace dataset~\cite{yi2014learning}, which originally has $494,414$ face images of $10,575$ identities. The first version was made by adding facial masks occlusions generated by MaskTheFace method~\cite{anwar2020masked} to every image of the dataset, and the second was generated by adding random occlusions in the forehead and periocular regions with one large black box to all images. The original dataset was augmented with the occluded versions, shuffled and splited into train, validation and test sets by a ratio of $0.7/0.2/0.1$ respectively. The training data was used to finetune the mask invariant network proposed by~\cite{huber2021mask}, named MaskInvHg, by freezing around $75\%$ of its learnable parameters, with the objective that he learns in a end2end manner to be invariant of occlusions in general. The cross entropy loss was used to train a facial classifier, which is a fully connected layer that receives as input the 512-d feature vector generated by the network and outputs the predicted face of the training dataset. The model was trained for $50$ epochs with a batch size of $64$, using the SGD optimizer, with a learning rate of $0.001$, and finished the training process with around 50\% accuracy on the validation dataset. Lastly, the fully connected layer was removed and the feature vector used as its input is the final output of the network. The model AFOIRNet\_1 was trained with the full synthetic augmented dataset, and the AFOIRNet\_2 was trained on a smaller version of the augmented dataset. The models AFOIRNet\_1 and AFOIRNet\_2 have $18269004$ and $14765727$ learnable parameters and size 283MB, 269MB, respectively. }


\begin{table*}[!h]
 \caption{Face verification results on three different protocols: protocol 3, protocol 4 and protocol 5. All protocols in the BLR-OP setting. The values of EER, FMR100, FMR1000 are displayed in percentage (\%). The rank orders the solutions from the best to the worst within that protocol. }
\label{table_results_2}
\centering
\resizebox{\textwidth}{!}{
\begin{tabular}{|l|nnncccc|cccc|cccc|}
\hline
  & \multicolumn{7}{|c}{Protocol 3} & \multicolumn{4}{|c}{Protocol 4} &\multicolumn{4}{|c|}{Protocol 5}    \\ 
\cmidrule{1-16} Method & GMean & IMean & AUC & EER & FMR100 & FMR1000 & Rank& EER & FMR100 & FMR1000 & Rank &  EER & FMR100 & FMR1000 & Rank\\ 
\hline
\hline
 Baseline UB & 0.417 & 0.003 & 0.960 & 9.233 & 13.567 & 16.300& -&1.967 & 2.800 & 6.600& -& 28.633 & 59.867 & 68.533& -\\
 \hline
 Baseline & 0.271 & 0.023 & 0.884 & 18.833 & 35.867 & 45.567& -& 11.500 & 30.333 & 40.467& - & 38.200 & 84.200 & 93.633& -\\
  \hline
  \hline
 SMT-OCFR1 & 0.488 & 0.746 & 0.879 & 20.800 & 36.267& 45.267 & 3& 14.967 & 30.767 & 38.933& 3& 45.200 & 94.433 & 96.933& 4\\
 \hline
 SMT-OCFR2 & 0.497 & 0.773 & 0.880 & 19.400 & 38.733 & 50.600& 4& 14.633 & 31.833 & 49.567& 4& 41.967 & 93.467 & 96.900& 3\\
 \hline
 AFOIRNet\_1 & 0.686 & 0.631 & 0.727 & 34.833 & 80.700 & 90.000& 5 & 31.767 & 78.333 & 89.367& 5& 41.600 & 94.700 & 98.967& 5\\
 \hline
 AFOIRNet\_2 & 0.650 & 0.599 & 0.699 & 35.800 & 81.267 & 91.233& 6& 32.833 & 80.233 & 91.900& 6& 42.533 & 95.367 & 99.400& 6\\
 \hline
 AdaFace4M & 0.303 & 0.015 & 0.938 & 12.633 & 20.833 & 33.033& 2& 6.000 & 12.233 & 23.766& 2& 32.133 & 72.100 & 88.567& 1\\
 \hline
 AdaFace12M & 0.294 & 0.011 & 0.940 & 11.933 & 20.333 & 27.800& 1 & 5.933 & 11.967 & 20.833& 1 & 33.000 & 74.133 & 89.267& 2\\

\hline
\end{tabular}
}
\end{table*}

\begin{tiny}
\begin{table*}[!h]
 \caption{Face verification results on two different protocols: protocol 6 and protocol 7. All protocols in the BLR-OP setting. The values of EER, FMR100, FMR1000 are displayed in percentage (\%). The rank orders the solutions from the best to the worst within that protocol. Includes information regarding the compactness of the models in terms of the number of learnable parameters and GFLOPs.}
\label{table_results_3}
\centering
\resizebox{\textwidth}{!}{
\begin{tabular}{|l|nnncccc|cccc|cc|}
\hline
  & \multicolumn{7}{|c}{Protocol 6} & \multicolumn{4}{|c}{Protocol 7} &\multicolumn{2}{|c|}{Compactness}    \\ 
\cmidrule{1-14} Method & GMean & IMean & AUC & EER & FMR100 & FMR1000 & Rank& EER & FMR100 & FMR1000 & Rank &  \# of learnable parameters & GFLOPs\\ 
\hline
\hline
 Baseline UB & 0.297 & 0.002 & 0.897 & 18.267 & 30.400 & 37.600& -& 14.100 & 22.633 & 26.600& -& 65 155 648 & -\\
 \hline
 Baseline & 0.182 & 0.010 & 0.793 & 28.867 & 55.133 & 59.667& -& 23.533 & 46.400 & 61.433& -& 65 155 648 & -\\
  \hline
  \hline
 SMT-OCFR1 & 0.575 & 0.762 & 0.779 & 30.167 & 57.567 & 67.100& 3& 26.100 & 48.467 & 57.800& 3& 131 206 393 & 24.35\\
 \hline
 SMT-OCFR2 & 0.565 & 0.739 & 0.784 & 29.200 & 59.467 & 69.600& 4& 25.167 & 49.933 & 60.300& 4& 65 131 000 & 24.20\\
 \hline
 AFOIRNet\_1 & 0.664 & 0.624 & 0.679 & 37.600 & 85.833 & 95.300& 5 & 37.067 & 81.900 & 89.933& 5& 65 150 912 & 12.09\\
 \hline
 AFOIRNet\_2 & 0.623 & 0.585 & 0.655 & 39.100 & 86.533 & 94.633& 6& 38.333 & 83.567 & 90.800& 6& 65 150 912& 12.09\\
 \hline
 AdaFace4M & 0.213 &  0.013 & 0.870 & 21.300 & 41.867 & 54.667& 2& 17.667 & 31.633 & 47.667& 2& 65 150 912&  12.09\\
 \hline
 AdaFace12M & 0.203 & 0.010 & 0.869 & 21.433 & 41.000 & 52.367 & 1 & 18.067 & 30.967 & 44.533& 1& 65 150 912&  12.09\\

\hline
\end{tabular}
}
\end{table*}
\end{tiny}

\section{Results and discussion}
\label{sec:results}

{In this section we present and discuss the results of the six solutions submitted to the competition and compare them with the results of the baseline approaches. The ranking results are spread across 7 different protocols and the final ranking is reported at the end of the section. AdaFace12M consistently beats the remaining submissions. Hence, it is considered the top ranked solution. Further, we show the results for Occluded-Occluded face verification on two protocols, the least difficult and the hardest one. }

\begin{figure*}[h!]
    \centering
    \begin{subfigure}[b]{0.49\linewidth}
       \includegraphics[height=5cm,width = \linewidth]{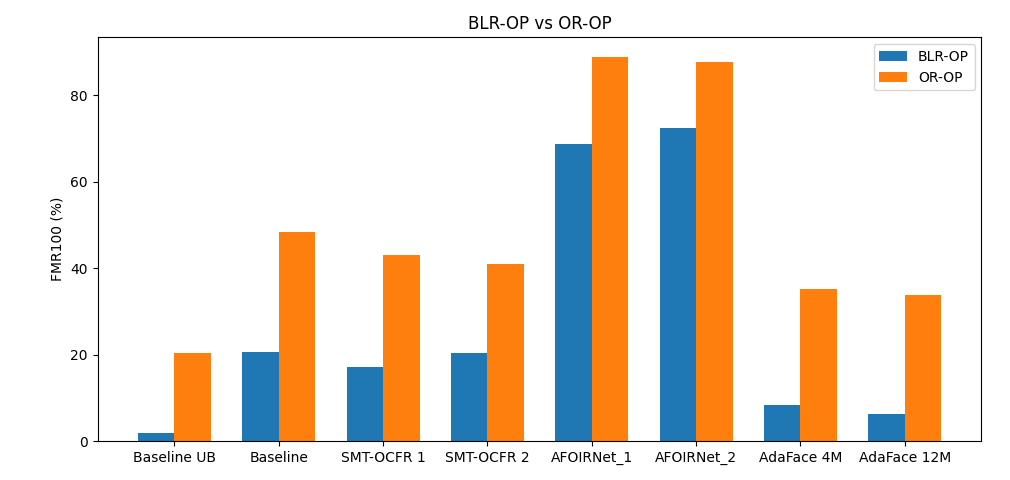}
       \caption{Protocol 1}
       \label{protocol1}
  \end{subfigure}
  \begin{subfigure}[b]{0.49\linewidth}
       \includegraphics[height=5cm, width = \linewidth]{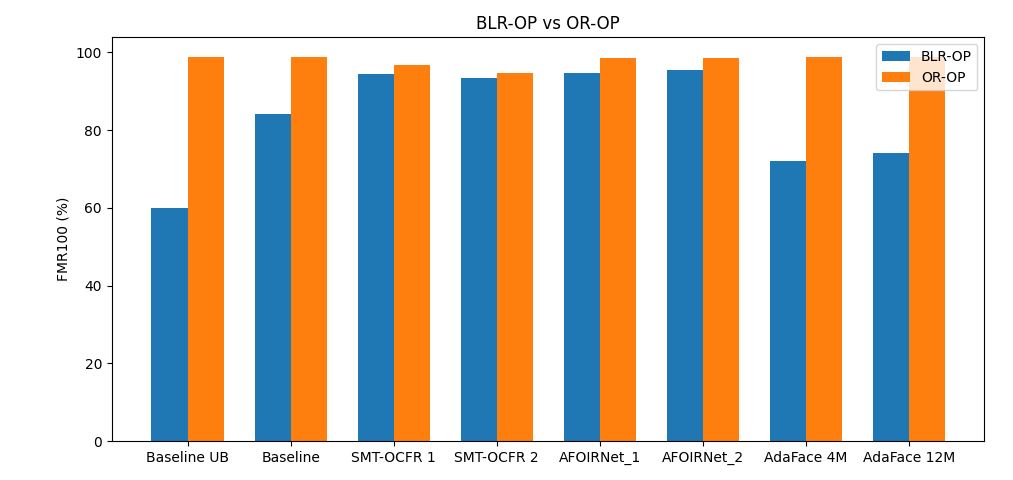}
       \caption{Protocol 5}
       \label{protocol5}
  \end{subfigure}

  \caption{Comparison of the performance of the Baseline algorithms and the submitted solutions on the easiest and the hardest Protocols, 1 and 5 respectively. Side by side analysis of the performance of BLR-OP and OR-OP. The algorithms are displayed in the following order "Baseline UB", "Baseline", "SMT-OCFR 1", "SMT-OCFR 2", "AFOIRNet\_1", "AFOIRNet\_2", "AdaFace 4M", "AdaFace 12M".}
  \label{fig:ocluded_to_occluded} 
\end{figure*}

\subsection{Non-occluded vs Occluded}

{The major evaluation of this competition relies on a face verification setting where the reference is not occluded and the probe is occluded. To understand the implication of small and single occlusions, it is first necessary to understand how the solutions perform on the non-occluded dataset (Protocol 0). Table~\ref{results_table_1} presents the results for the first three protocols (0,1,2). It is interesting to see that four of the six solutions outperform the Baseline (not UB). Moreover, SMT-OCFR1 achieves better FMR100 results than the Baseline UB, which is not the case with EER and FMR1000. It is not surprising that SMT-OCFR1 and SMT-OCFR2 are the only two solutions to achieve comparable results to the Baseline UB since they are the only solutions to align the input images. Surprisingly, in Protocol 1, a drop in the alignment success rate - measured from the alignment success flag output of the solutions - of just 3.1 percentage points together with simple occlusions is enough to create a significant performance gap between SMT-OCFR1 and the Baseline UB. Nonetheless, the number of solutions that remain superior to the Baseline (not UB) does not change. AdaFace4M and AdaFace12M have shown impressive robustness to occlusions, which is visible in the FMR100 value being at least 2.4x smaller. The performance degrades rather quickly when we evaluate in a protocol composed exclusively of a combination of two occlusions as seen in the performance of Protocol 2. While Baseline UB shows that having knowledge of the landmarks strongly benefits the performance, SMT-OCFR1 and SMT-OCFR2 show that failing to align the images might be more harmful to the performance than not using any alignment at all like AdaFace4M and AdaFace12M. These latter models show impressive results that are significantly better than the Baseline (without alignment), although they use alignment. }

{The performance on Protocol 2 is mitigated when there is no constraint to only use two joint occlusions. For instance, Protocol 3, shown in Table~\ref{table_results_2} shows a reduction in all the errors for all the methods in comparison to Protocol 2 in Table~\ref{results_table_1}. Nonetheless, the landscape remains intact as only AdaFace4M and AdaFace12M surpass the Baseline, but not the Baseline UB. If we create a restriction that Protocol 3 cannot have simultaneous upper (U) and lower (L) face occlusions, which hide the majority of the face, then we obtain Protocol 4 also seen in Table~\ref{table_results_2}. The results are still far from the ones achieved with single occlusions in Protocol 1, which shows that despite being a factor for a large portion of the errors, the combination of U+L occlusions is not a single-handedly responsible for the performance degradation, since  E+L, T+L and E+T consist of a significant source of error to the models. Protocol 5 is virtually the most difficult since it only allows triple occlusions. The gap between the Baseline, SMT-OCFR1, and SMT-OCFR2 increases as these submitted solutions fall short to have an FMR100 smaller than  90\%, which might be caused by attempts to align the input images with a success rate of 61.2\%. On the other hand, the performance of AdaFace4M and AdaFace12M, which work with unaligned samples shines and keeps surpassing the Baseline. }

{Table~\ref{table_results_3} shows the results on Protocols 6 and 7, which are relaxations of Protocol 5 to include double and single occlusions, respectively. The overall performance remains as expected from previous results. It is also possible to observe the number of parameters and GFLOPs of the models. Moreover, while it is not stated explicitly, SMT-OCFR1 and SMT-OCFR2 require two extraordinary steps of preprocessing, which also increase their computational complexity. }

\begin{tiny}
\begin{table}[!h]
\caption{Final ranking of the solutions submitted. Solutions in bold surpassed the performance of the baseline (without access to the landmarks) in at least one protocol. }
\label{table_final_ranking}
\centering
\begin{tabular}{|l|c|c|}
\hline
Method   & Avg. rank (7 protocols) & Rank\\ 
\hline
 \textbf{AdaFace12M} &  1.14 & 1\\
 \hline
 \textbf{AdaFace4M} & 1.85 &  2\\
 \hline
 \textbf{SMT-OCFR1} & 3.29 & 3\\
 \hline
 \textbf{SMT-OCFR2} & 3.71 & 4\\
 \hline
 AFOIRNet\_1 & 5.00 & 5\\
 \hline
 AFOIRNet\_2 & 6.00 & 6\\
\hline
\end{tabular}
\end{table}
\end{tiny}

{The solutions are ranked accordingly to the average of performance on the seven protocols. The final ranking is shown in Table~\ref{table_final_ranking}, as expected from the individual analysis of each protocol AdaFace12M takes the lead closely followed by AdaFace4M. This highlights the lower performance of algorithms that strongly rely on alignment for occluded FR. In a sense, this result creates new grounds for further research on which situations are beneficial to discard the alignment step.}



\subsection{Occluded vs Occluded}

We further extended our analysis to occluded reference versus occluded probe recognition. The main goal of this analysis is to understand how these algorithms behave when exposed to stronger occlusions on both sides of the comparison. Figure~\ref{fig:ocluded_to_occluded} shows the performance of the algorithm in Protocol 1 (Figure~\ref{protocol1}) and Protocol 2 (Figure~\ref{protocol5}), which are the less and the more occluded, respectively. It is also interesting to note that the SMT-OCFR alignment success rate decreases to 22.2\% in Protocol 5 when both the probe and the reference are occluded. And while in Protocol 1, the best performing algorithms retain their advantage, the strong occlusions present in Protocol 5 redefine the order of the best performing solutions. All the solutions achieve results above the 90\% FMR100, which is considered inadequate. Nonetheless, it is worth noting that this is a demanding scenario, and a significant portion of the pairs might not have sufficient information available to support a valid verification. 

\section{Conclusion}
\label{sec:conclusion}
IJCB-OCFR-2022 is organised to motivate novel research towards enhancing OFR accuracy.  Three participating teams have registered in the competition and submitted solutions. Each of them submitted two valid solutions with total of six valid solutions. The submitted solutions are evaluated on a novel synthetically occlusion benchmark, namely the OCFR2022, created by the competition organisers. Additionally, extensive experimental evaluations are conducted based on seven evaluation protocols designed by the organisers to simulate different occlusion scenarios.
Some of the submitted solutions showed resilience to the occlusion effect and scored competitive performances.

\subsection*{Acknowledgments}
This work was financed by National Funds through the Portuguese funding agency, FCT - Fundação para a Ciência e a Tecnologia within project LA/P/0063/2020, and within the PhD grants ``2021.06872.BD'' and ``SFRH/BD/137720/2018''. This research work has been also partially funded by the German Federal Ministry of Education and Research and the Hessen State Ministry for Higher Education, Research and the Arts within their joint support of the National Research Center for Applied Cybersecurity ATHENE.


{\small
\bibliographystyle{ieee}
\bibliography{egbib}
}

\end{document}